\documentclass[letterpaper, 10 pt, conference]{ieeeconf}  

\usepackage{amsmath}
\usepackage{amssymb}
\usepackage{algorithm}
\usepackage{algpseudocode}
\usepackage{graphicx}
\usepackage{cite}
\IEEEoverridecommandlockouts                              

\title{\LARGE \bf
PAMAE: Phase-Aware-MoE Action Experts Towards Reliable Flow-Matching Vision-Language-Action Policies
}

\author{Jiayu Yang, Tao Yang, Xiang Chang, Fei Chao, Changjing Shang, and Qiang Shen%
\thanks{J. Yang, T. Yang, and F. Chao are with the Department of Artificial Intelligence, School of Informatics, Xiamen University, Xiamen, China. X. Chang, C. Shang, and Q. Shen are with the Department of Computer Science, Aberystwyth University, Aberystwyth, U.K. This work was supported by the 
Xiamen Municipal Natural Science Foundation Project (No.~3502Z202573010) 
and the Key Program of the National Natural Science Foundation of 
China Joint Fund (No.~U23A20383).}%
}

\begin{document}

\maketitle
\thispagestyle{empty}
\pagestyle{empty}

\begin{abstract}

Reliable action generation for multi-stage robotic manipulation remains challenging for Vision-Language-Action (VLA) models. While existing flow-matching VLA policies offer strong multimodal grounding and generalization, they typically employ a single shared action expert, limiting their ability to capture phase-specific control patterns across distinct execution stages. We propose a plug-and-play Phase-Aware Mixture-of-Experts Action Module (PAMAE), as a step towards more reliable phase-consistent action generation. PAMAE replaces the original flow-matching action expert with a sparse expert mixture while preserving the pretrained VLA backbone. PAMAE introduces a phase-aware router that leverages execution-phase cues to allocate action generation across experts, supported by a lightweight phase prediction head and a routing alignment objective. To stabilize specialization, we adopt a two-stage training scheme that first warms up the expert module under the standard flow-matching loss and then optimizes phase-consistent routing under auxiliary supervision. On multi-stage manipulation simulation tasks, PAMAE improves task success by up to \textbf{9.2\%} over strong VLA baselines. Further ablations show that both phase-supervised routing and staged optimization are essential for the observed gains. Our results highlight phase-consistent expert allocation as an effective mechanism for improving the reliability and action quality of flow-matching VLA policies.

\end{abstract}

\section{INTRODUCTION}

Reliable action generation for multi-stage robotic manipulation remains a central challenge in embodied intelligence \cite{ref1,ref2,ref3}. Vision-Language-Action (VLA) models have recently emerged as a promising paradigm for general-purpose robot control, unifying visual perception, language understanding, and action generation within a single policy \cite{ref4,ref5,ref6}. Benefiting from large-scale pretraining, VLA models capture rich semantic priors from large and diverse datasets, which endow them with strong multimodal grounding, flexible task specification, and broad generalization across a wide spectrum of robotic manipulation tasks \cite{ref7,ref8}. These advantages make VLA policies a particularly attractive foundation for multi-stage manipulation, although reliably adapting them to temporally structured execution remains an open problem \cite{ref9}.

One key challenge lies in the reliance on a single shared action expert in most flow-matching VLA systems \cite{ref10,ref11}. Representative models such as $\pi_0$ \cite{ref12} and $\pi_{0.5}$ \cite{ref13} benefit from strong multimodal grounding and continuous action generation, yet they typically rely on a single expert to predict control signals across all stages of task execution. This design is suboptimal for multi-stage manipulation, where different phases often exhibit distinct control patterns, transition dynamics, and precision requirements. For instance, approach, contact, and transport stages may require substantially different velocity fields and action sensitivities, yet they are all modeled under a unified parameterization. Consequently, phase-specific behaviors can be blurred, making the policy less responsive to stage-dependent control demands. This issue is especially pronounced in long-horizon and contact-rich tasks, where inadequate modeling of phase-specific action patterns can reduce both action precision and task robustness \cite{ref14,ref15}.

Another challenge is that existing VLA with MoE methods are not explicitly designed to exploit execution-phase structure in low-level action generation. Most prior approaches organize experts around task semantics, scaling objectives, embodiment differences, or auxiliary modalities, even though in flow-matching manipulation policies, low-level control variation often arises across execution phases \cite{ref16,ref17,ref18,ref19}. Different tasks may share similar control behaviors within the same phase, whereas different phases of a single task can require substantially different control patterns and transition dynamics \cite{ref20}. Consequently, even when expert mixtures are introduced, expert allocation may remain poorly aligned with phase-specific action structure. This exposes a key limitation for flow-matching VLA models: effective control requires not only greater expert capacity, but also phase-consistent allocation of low-level action experts \cite{ref21}.

To address these limitations, we propose a plug-and-play framework for phase-aware expert allocation in pretrained flow-matching VLA policies, PAMAE. PAMAE preserves the pretrained VLA backbone while replacing the original shared action expert with a sparse mixture of specialized action experts. A phase-aware router, supported by a lightweight phase prediction head and a routing alignment objective, allocates action generation according to execution-phase structure, enabling low-level actions to better capture phase-dependent control patterns and transition dynamics in multi-stage manipulation.

PAMAE is trained with a two-stage optimization strategy. The action experts are first warmed up under the standard flow-matching objective, after which phase-consistent routing supervision is introduced to encourage structured expert specialization. In late training, auxiliary routing constraints are gradually relaxed to account for phase ambiguity and shift optimization toward action quality refinement. Together, these designs provide a stable way to improve phase-consistent expert allocation and action generation in pretrained flow-matching VLA policies.

The main contributions of this paper are summarized as follows:

\textbf{(1) PAMAE: A phase-aware MoE action expert for flow-matching VLA policies.}
We introduce PAMAE, a plug-and-play framework that preserves the pretrained VLA backbone while replacing the shared action expert with a sparse mixture of specialized experts, enabling low-level action generation to better capture phase-dependent control patterns in multi-stage manipulation.

\textbf{(2) Phase-supervised routing with stable two-stage optimization.}
We propose a routing mechanism where weak phase signals guide expert allocation rather than directly supervising action outputs. To stabilize specialization, we adopt a two-stage scheme with expert warm-up, phase-consistent routing supervision, and later relaxation.

\textbf{(3) Phase-consistent expert allocation without inference-time phase labels.}
PAMAE learns phase-consistent expert allocation without explicit phase labels at inference time, improving task success by up to \textbf{9.2\%} over strong flow-matching VLA baselines and enabling phase-aware control without additional inference-time supervision.

\section{BACKGROUND}

\subsubsection{Vision-Language-Action Models}

VLA models unify visual perception, language understanding, and robot control by mapping multimodal observations and task instructions to executable actions \cite{ref4,ref5,ref6}. Built on large vision-language backbones, recent VLA systems, such as $\pi_0$ and $\pi_{0.5}$ \cite{ref12,ref13}, demonstrate strong semantic grounding, broad task generalization, and flexible instruction following across diverse manipulation settings. In practice, VLA policies usually use autoregressive, chunked, diffusion-based, or flow-matching action generation strategies to generate actions, depending on the desired balance between expressivity, temporal coherence, and control frequency \cite{ref4,ref22,ref23,ref12}. In flow-matching VLA policies, the action generator learns a conditional velocity field $v_{\theta}(\mathbf{z}_{\tau}, h)$, where $\mathbf{z}_{\tau}$ denotes an intermediate action state at flow time $\tau$, and $h$ is the multimodal context encoded from observations, instructions, and robot states \cite{ref12}. This formulation is well-suited to continuous high-frequency action generation and therefore provides a strong foundation for dexterous and long-horizon manipulation.

\subsubsection{Mixture-of-Experts Models}

MoE architectures increase model capacity and specialization by routing different inputs to different experts \cite{ref24,ref25}. In general, an MoE module combines multiple experts through routing weights $y = \sum_{i=1}^{M} g_i(x)\, f_i(x)$, where $f_i$ denotes the $i$-th expert and $g_i(x)$ is produced by a routing network, often with sparse top-$k$ activation. In robotics and VLA systems, MoE-based designs have been explored for model scaling, embodiment heterogeneity, skill specialization, and multimodal fusion. However, most existing MoE designs for robotics and VLA systems are not explicitly structured around execution phases in low-level manipulation. This is particularly relevant for flow-matching VLA policies, where reliable action generation depends not only on expressive expert capacity, but also on routing mechanisms that can adapt to phase-dependent control structure \cite{ref21}.

\section{METHOD}

\begin{figure*}[t]
    \centering
    \includegraphics[width=.98\textwidth]{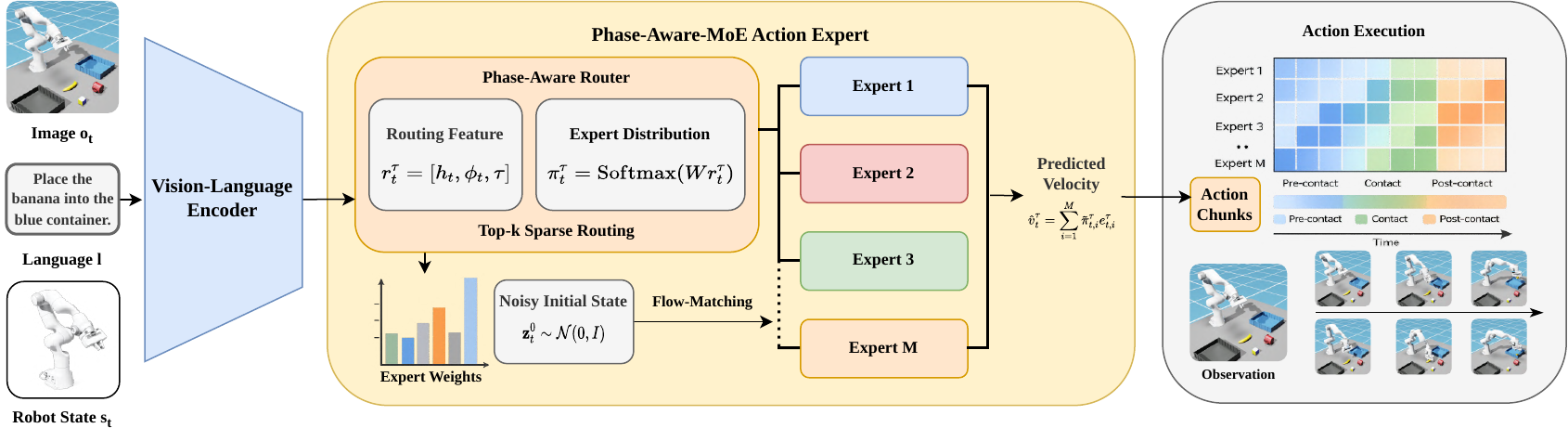}
    \caption{
    Overview of PAMAE during execution. Given the current observation \(o_t\), language instruction \(l\), and robot state \(s_t\), the pretrained vision-language encoder produces the context \(h_t\). PAMAE replaces the original action expert with a phase-aware sparse MoE expert: the router constructs \(r_t^\tau=[h_t,\phi_t,\tau]\), predicts top-\(k\) sparse expert weights, and aggregates expert outputs to produce the velocity prediction \(\hat v_t^\tau\), which is then integrated to generate the action chunk. The expert-usage heatmap illustrates the intended phase-related routing, with different experts dominating in different execution phases.
    }
    \label{fig:overview}
\end{figure*}

As illustrated in Fig.~\ref{fig:overview}, we propose PAMAE, a phase-aware action generation framework that enhances pretrained VLA policies for reliable multi-stage robotic manipulation. Instead of using a single shared action expert across all execution stages, PAMAE replaces the original action module with a sparse mixture of specialized action experts coordinated by a phase-aware router. During training, weak phase supervision guides expert allocation and induces phase-consistent specialization, while leaving the pretrained VLA backbone unchanged. At inference time, the learned router autonomously dispatches action generation without requiring explicit phase labels, preserving the original flow-matching generation process while improving action precisionfor autonomous execution in temporally structured manipulation tasks.

\subsection{Preliminary}

We consider a robot demonstration dataset $\mathcal D=\left\{\{(o_t,s_t,a_t,l)\}_{t=1}^{T_n}\right\}_{n=1}^{N}$, where \(o_t\) denotes visual observations, \(s_t\) denotes robot states, \(a_t\) denotes continuous control actions, and \(l\) is the language instruction. Following flow-matching VLA policies, the objective is to predict a future action chunk $\mathbf a_t=[a_t,a_{t+1},\dots,a_{t+H-1}]$, where $H$ denotes the action horizon.

Given the multimodal inputs, a pretrained VLA backbone encodes the current context as $h_t=f_{\mathrm{enc}}(o_t,l,s_t)$. Instead of directly regressing $\mathbf a_t$, flow matching models action generation through intermediate noisy action states. Specifically, given noise $\epsilon\sim\mathcal N(0,I)$ with the same shape as $\mathbf a_t$, and a flow time $\tau\in[0,1]$, we define the interpolated action state as $\mathbf z_t^\tau=\tau \mathbf a_t+(1-\tau)\epsilon$. A standard flow-matching action expert then predicts a conditional velocity field $v_\theta(\mathbf z_t^\tau, h_t)$, which is trained to match the target velocity $u^\star(\mathbf z_t^\tau\mid \mathbf a_t)=\epsilon-\mathbf a_t$. The corresponding flow-matching objective is $\mathcal L_{\mathrm{fm}}
=
\mathbb E
\left[
\left\|
v_\theta(\mathbf z_t^\tau, h_t)-(\epsilon-\mathbf a_t)
\right\|_2^2
\right]$ \cite{ref12}.
This formulation serves as the action-generation basis for our method. Building on it, we reformulate the original shared action expert as a phase-aware sparse mixture of experts for multi-stage manipulation.

\subsection{PAMAE: Phase-Aware-MoE Action Expert}

Standard flow-matching VLA policies parameterize low-level action generation with a single shared action expert that predicts a conditional velocity field $v_\theta(\mathbf z_t^\tau, h_t)$, where $h_t$ denotes the multimodal context encoded by the pretrained VLA backbone. PAMAE replaces this shared action expert with a sparse mixture of phase-aware action experts while preserving both the pretrained VLA backbone and the original flow-matching generation interface.

Given the current observation, language instruction, and robot state, the backbone produces a context representation $h_t=f_{\mathrm{enc}}(o_t,l,s_t)$. To condition action generation on both the noisy action state and the multimodal context, we introduce a shared action trunk $q_t^\tau=f_{\mathrm{shared}}(\mathbf z_t^\tau,h_t)$, which extracts a common latent representation for all experts. On top of this shared trunk, we instantiate \(M\) action experts, $e_{t,i}^\tau=f_i^{\mathrm{expert}}(q_t^\tau), i=1,\dots,M$, where each expert predicts a candidate velocity-field component for the current noisy action state.

To allocate action generation across experts, PAMAE introduces a phase-aware router. The router takes as input a routing feature $r_t^\tau=[h_t,\phi_t,\tau]$, where \(\phi_t\) is a low-dimensional execution descriptor constructed from lightweight trajectory signals. In particular, we define $\phi_t=[g_t,\Delta g_t,\|a_{t-1}\|_2,\rho_t]$, where \(g_t\) denotes the gripper state, \(\Delta g_t=g_t-g_{t-1}\) captures gripper variation, \(\|a_{t-1}\|_2\) measures recent action magnitude, and \(\rho_t\) is a coarse normalized progress cue. In practice, we set $\rho_t=t/T$, where \(T\) is the 95th percentile of trajectory lengths in the training set, and clip \(\rho_t\) at 1 for timesteps beyond \(T\). These signals provide lightweight cues about the current execution status while remaining fully available at inference time. The router outputs a distribution over experts $\pi_t^\tau=\mathrm{Softmax}(Wr_t^\tau)$, and sparse top-\(k\) routing is applied to obtain \(\tilde\pi_t^\tau\). The final velocity prediction is computed as the weighted aggregation of expert outputs: $\hat v_t^\tau=\sum_{i=1}^{M}\tilde\pi_{t,i}^\tau e_{t,i}^\tau$.

This MoE predictor directly replaces the original shared action expert in the base flow-matching VLA policy, enabling low-level action generation to be organized through phase-aware sparse expert allocation while preserving the original backbone and flow-based generation process. In practice, replacing the shared action expert with PAMAE increases the total parameter count from 300\,M to 450\,M (using $\pi_0$ backbone).

\subsection{Training Pipeline}

PAMAE is trained under the original flow-matching objective, augmented with phase-aware auxiliary losses that shape router behavior and stabilize expert specialization. During training, each timestep is associated with a pseudo phase label $y_t^{\mathrm{phase}} \in \{0,1,2\}$, corresponding to pre-contact, contact/manipulation, and post-contact stages, respectively.

We adopt a simple rule-based phase definition based on gripper motion and end-effector movement. Let \(g_t\) denote the gripper state and \(p_t\) denote the end-effector position, and define \(\Delta g_t = g_t-g_{t-1}\), \(v_t=\|p_t-p_{t-1}\|_2\). We then define the contact onset and transport onset as \(t_c=\min\{t \mid \Delta g_t<-\delta_g\}\), \(t_m=\min\{t>t_c \mid v_t>\delta_v \ \text{and}\ g_t<\delta_c\}\), where \(\delta_g\), \(\delta_v\), and \(\delta_c\) are fixed thresholds for gripper closing, end-effector motion, and closed-gripper state, respectively. The phase label is then assigned as
\begin{equation}
y_t^{\mathrm{phase}}=
\begin{cases}
0, & t<t_c,\\
1, & t_c\le t<t_m,\\
2, & t\ge t_m.
\end{cases}
\end{equation}

This three-way partition should be understood as a coarse phase abstraction. It does not assume that a manipulation trajectory contains only a single pre-contact, contact, and post-contact segment. Rather, complex tasks may include multiple sub-stages with similar control characteristics, which are grouped under the same coarse phase category for routing supervision.

Given the pseudo phase label, PAMAE is trained with the standard flow-matching objective
\begin{equation}
\mathcal L_{\mathrm{fm}}
=
\mathbb E
\left[
\left\|
\hat v_t^\tau-u^\star(\mathbf z_t^\tau\mid \mathbf a_t)
\right\|_2^2
\right],
\end{equation}
together with three auxiliary objectives and a lightweight load-balancing regularizer. First, we introduce a lightweight phase prediction head operating on the same routing feature as the router. Concretely, the phase head is implemented as a small MLP $p_t^{\mathrm{phase}}=\mathrm{Softmax}\!\left(\mathrm{MLP}_{\mathrm{phase}}(r_t^\tau)\right)$,
which predicts the probability of each coarse execution phase. The phase supervision loss is defined as
\begin{equation}
\mathcal L_{\mathrm{phase}}
=
\mathrm{CE}(p_t^{\mathrm{phase}}, y_t^{\mathrm{phase}}),
\end{equation}
where \(y_t^{\mathrm{phase}}\) is the pseudo phase label. This auxiliary head is used only during training to encourage the routing feature to retain explicit phase information, and is not required at inference time. Second, we impose a route alignment objective
\begin{equation}
\mathcal L_{\mathrm{route}}
=
\mathrm{KL}(\pi_t^\tau \,\|\, p_t^{\mathrm{target}}),
\end{equation}
where \(p_t^{\mathrm{target}} = p^{(y_t^{\mathrm{phase}})}\) is the routing prior associated with the current phase label \(y_t^{\mathrm{phase}}\), and \(p^{(0)},p^{(1)},p^{(2)} \in \mathbb{R}^M\) denote phase-specific target distributions over the \(M\) experts. Third, to reduce routing jitter across neighboring timesteps, we apply
\begin{equation}
\mathcal L_{\mathrm{smooth}}
=
\sum_t \|\pi_t^\tau-\pi_{t-1}^\tau\|_2^2.
\end{equation}
To prevent expert collapse, we further introduce a lightweight load-balancing regularizer on batch-level expert utilization. Let $\bar{\pi}_i=\frac{1}{B}\sum_{b=1}^{B}\pi_{b,i}$, where \(B\) is the batch size and \(\pi_{b,i}\) is the routing probability assigned to expert \(i\) for sample \(b\). Let \(u=[1/M,\dots,1/M]\) denote the uniform expert distribution. We define
\begin{equation}
\mathcal L_{\mathrm{lb}}
=
\mathrm{KL}(\bar{\pi}\,\|\,u),
\end{equation}
where \(\bar{\pi}=[\bar{\pi}_1,\dots,\bar{\pi}_M]\) and \(M\) is the number of experts.

We optimize PAMAE in two stages to stabilize expert specialization while avoiding premature routing collapse. In Stage~1, we warm up the sparse action experts using only the main flow-matching objective together with a very small load-balancing term:
\begin{equation}
\mathcal L^{(1)}
=
\mathcal L_{\mathrm{fm}}
+\lambda_b^{(1)}\mathcal L_{\mathrm{lb}},
\end{equation}
where \(\lambda_b^{(1)}\) is set to a very small value. In this stage, \(\mathcal L_{\mathrm{lb}}\) serves only as a weak anti-collapse regularizer rather than an objective for uniform expert usage, which could otherwise hinder later specialization. In Stage~2, we introduce phase-supervised routing objectives to encourage structured expert specialization:
\begin{equation}
\mathcal L^{(2)}
=
\mathcal L_{\mathrm{fm}}
+\lambda_1\mathcal L_{\mathrm{phase}}
+\lambda_2\mathcal L_{\mathrm{route}}
+\lambda_3\mathcal L_{\mathrm{smooth}}
+\lambda_4\mathcal L_{\mathrm{lb}}.
\end{equation}
Here, \(\mathcal L_{\mathrm{phase}}\) encourages the routing feature to retain explicit phase information, \(\mathcal L_{\mathrm{route}}\) aligns expert allocation with phase-conditioned routing targets, and \(\mathcal L_{\mathrm{smooth}}\) suppresses routing jitter across neighboring timesteps. To account for the ambiguity and imbalance of real-world execution phases, we gradually relax the auxiliary routing losses in late training. Specifically, during the final 30\% of Stage~2, we linearly anneal the weights of \(\mathcal L_{\mathrm{phase}}\), \(\mathcal L_{\mathrm{route}}\), and \(\mathcal L_{\mathrm{lb}}\) to 10\% of their initial values, while keeping \(\mathcal L_{\mathrm{fm}}\) and \(\mathcal L_{\mathrm{smooth}}\) unchanged. This allows optimization to shift from enforcing expert separation toward refining action quality under the main flow-matching objective, while preserving weak structural regularization for stable routing.

\subsection{Execution}

At inference time, PAMAE preserves the original flow-matching generation process and does not require explicit phase labels. Given the current observation, language instruction, and robot state, the pretrained VLA backbone produces the multimodal context \(h_t=f_{\mathrm{enc}}(o_t,l,s_t)\). Starting from an initial noisy action state \(\mathbf z_t^{0}\sim \mathcal N(0,I)\), the policy iteratively refines the action chunk. At each flow time \(\tau\), the router uses $r_t^\tau=[h_t,\phi_t,\tau]$ to predict sparse expert weights \(\pi_t^\tau\), and the top-\(k\) experts are aggregated to form the velocity field \(\hat v_t^\tau\). The action state is then updated with the same flow integration rule as the base policy: $\mathbf z_t^{\tau+\Delta\tau}=\mathbf z_t^\tau+\Delta\tau\,\hat v_t^\tau$.

After a fixed number of integration steps, the final refined state is taken as the predicted action chunk \(\hat{\mathbf a}_t\). Therefore, PAMAE adds no phase supervision requirement at inference time while enabling phase-consistent sparse expert allocation for more robust multi-stage action generation.

\section{EXPERIMENTATION}

Our experiments evaluate whether phase-aware expert allocation improves flow-matching VLA policies for multi-stage manipulation. Specifically, we study whether PAMAE \textbf{(1)} improves task success over strong VLA baselines, \textbf{(2)} relies on phase-supervised routing and two-stage optimization for effective expert specialization, and \textbf{(3)} produces temporally coherent and phase-consistent expert allocation at inference time without explicit phase labels. All main results are obtained on simulated multi-stage manipulation tasks. Besides task success, we report average dominant-expert run length and phase-conditioned dominance purity (PCP) to analyze routing behavior and phase-consistent expert allocation.

\subsection{Experimental Setup}

We evaluate PAMAE on five simulated multi-stage manipulation tasks designed to test temporally structured execution, where different stages require distinct control patterns and transition dynamics. All methods are evaluated under the same setup, enabling a controlled assessment of whether phase-aware expert allocation improves low-level action generation in flow-matching VLA policies.

The tasks include: \textbf{Table-Cleaning (T.C.)} (shown in Fig.~{}), which requires clearing a cluttered tabletop and placing a specified target object into a designated container; \textbf{Drawer-Cycle (D.C.)}, where the robot aligns with a drawer handle, opens the drawer, places an object inside, and closes the drawer; \textbf{Lid-Open (L.O.)}, which involves opening a container lid, retrieving an object, and closing the lid afterward; \textbf{Shelf-Insert (S.I.)}, where the robot inserts an object into a constrained shelf slot; and \textbf{Cup-Upright (C.U.)}, where the robot reorients a tilted or overturned cup and places it stably at a target location. Together, these tasks cover diverse multi-stage behaviors, including approach, contact, transport, insertion, and release.

Each task is evaluated over 20 trials under 5 random seeds, yielding 100 runs per task, and success rates are computed over all runs. All compared methods use the same simulator, observation setting, action interface, and training configuration, with visual observations resized to \(256\times256\). Unless otherwise specified, each PAMAE variant uses the same backbone, action horizon, control frequency, and flow integration steps as its corresponding baseline, so that performance differences can be attributed to the proposed phase-aware MoE action expert rather than differences in model scale or experimental setup. All PAMAE variants use \(M=6\) experts with top-\(k=3\) sparse routing, and all experiments are conducted on \(4\times\)A100 GPUs.

\begin{table}[t]
\centering
\caption{Task success rates (\%) across five multi-stage manipulation simulation tasks: Table-Cleaning (T.C.), Drawer-Cycle (D.C.), Lid-Open (L.O.), Shelf-Insert (S.I.), and Cup-Upright (C.U.).}
\label{tab:main_results}
\begin{tabular}{lcccccc}
\hline
\textbf{Method} & \textbf{T.C.} & \textbf{D.C.} & \textbf{L.O.} & \textbf{S.I.} & \textbf{C.U.} & \textbf{Avg.} \\
\hline

$\pi_0$ & 77.0 & 75.0 & 72.0 & 64.0 & 81.0 & 73.8 \\
$\pi_0$ (labeled) & 78.0 & 75.0 & 70.0 & 66.0 & 84.0 & 74.6 \\
$\pi_{0.5}$ & 88.0 & 84.0 & 86.0 & 79.0 & 92.0 & 85.8 \\
ProgressVLA & 80.0 & 80.0 & 75.0 & 69.0 & 87.0 & 78.2 \\
PAMAE($\pi_0$, ours) & 85.0 & 81.0 & 84.0 & 75.0 & 90.0 & 83.0 \\
PAMAE($\pi_{0.5}$, ours) & \textbf{93.0} & \textbf{89.0} & \textbf{92.0} & \textbf{86.0} & \textbf{97.0} & \textbf{91.4} \\
\hline
\end{tabular}
\end{table}

\subsection{Main Results}

Table~\ref{tab:main_results} summarizes the results across five multi-stage manipulation tasks. PAMAE consistently improves task success over strong flow-matching VLA baselines. We compare with \(\pi_0\) \cite{ref12}, \(\pi_0\) (labeled), \(\pi_{0.5}\) \cite{ref13}, and ProgressVLA \cite{ref26}. Here, \(\pi_0\) (labeled) is a control baseline that uses the same coarse phase supervision as PAMAE during training, but keeps the original single shared action expert without MoE routing or expert specialization. At inference time, it follows the same procedure as raw \(\pi_0\) and requires no phase outputs. ProgressVLA is a progress-guided diffusion policy that uses a learned progress estimator and differentiable progress guidance for long-horizon manipulation.

We further instantiate our method on two representative backbones, denoted as PAMAE(\(\pi_0\)) and PAMAE(\(\pi_{0.5}\)), where only the original shared action expert is replaced, while the pretrained backbone, action horizon, and flow generation procedure remain unchanged. For controlled VLA-family comparisons, each PAMAE variant uses the same pretrained backbone and fine-tuning protocol as its corresponding raw policy, so that performance differences can be attributed to the proposed phase-aware-MoE action expert rather than backbone scale or additional task-specific retraining.

As shown in Table~\ref{tab:main_results}, PAMAE improves task success over its corresponding backbone models across all five tasks. The improvements are stable across repeated runs. In particular, PAMAE(\(\pi_0\)) improves the average task success rate from \textbf{73.8\%} to \textbf{83.0\%}, while PAMAE(\(\pi_{0.5}\)) improves it from \textbf{85.8\%} to \textbf{91.4\%}. Compared with $\pi_0 (labeled)$, these gains indicate that simply exposing phase information during training is insufficient; phase-supervised routing and expert specialization are both necessary to translate execution-phase structure into better low-level action generation. Compared with ProgressVLA, PAMAE further suggests that the observed gains are not explained by coarse progress cues alone, but instead arise from phase-consistent expert allocation within the action expert.

As a further analysis of the learned routing behavior, we examine whether PAMAE exhibits temporally coherent and phase-consistent expert allocation. For each action chunk, we define the dominant expert as the one with the largest routing weight after averaging over flow integration steps, and measure its average run length, i.e., the average number of consecutive action chunks controlled by the same dominant expert. A larger run length indicates that expert allocation is stable over time rather than switching erratically at every step. 

We also evaluate the dependence between routing and execution phase using the Phase-Conditioned Dominance Purity (PCP): $\mathrm{PCP}
=
\frac{1}{|\mathcal P|}
\sum_{p\in\mathcal P}
\max_i
P(d_t=i \mid y_t^{\mathrm{phase}}=p)$,
where \(d_t\) denotes the dominant expert at action chunk \(t\), \(y_t^{\mathrm{phase}}\) is the coarse pseudo phase label, and \(\mathcal P=\{0,1,2\}\) is the set of coarse phases. In our experiments, PAMAE($\pi_0$) achieves an average dominant-expert run length of 8 action chunks and a PCP of 89.0\%, indicating that its routing is both temporally coherent and strongly correlated with execution-phase structure.

These results show that replacing the original shared action expert with the proposed phase-aware sparse MoE module consistently improves task success in temporally structured manipulation. These gains suggest that the improvement comes not merely from additional supervision or coarse progress cues, but from phase-consistent expert allocation during low-level action generation.

\subsection{Ablation Studies}

We conduct ablation studies to isolate the contributions of \textbf{(1)} the proposed phase-supervised routing design and \textbf{(2)} the staged training scheme with expert warm-up. All variants use the same \(\pi_0\)-based backbone, training data, and evaluation protocol as the full PAMAE model to ensure a controlled comparison. Specifically, we compare the raw \(\pi_0\) backbone (Base), the full PAMAE model (Full), and three reduced variants: \textbf{-S1}, which removes Stage~1 warm-up and trains the full objective directly from the beginning; \textbf{-R}, which removes the routing alignment loss \(\mathcal L_{\mathrm{route}}\); and \textbf{-P}, which removes the auxiliary phase prediction loss \(\mathcal L_{\mathrm{phase}}\).

As shown in Table~\ref{tab:ablation}, the full model achieves the best overall performance. Compared with the raw \(\pi_0\) backbone, PAMAE improves the average task success rate from \textbf{73.8\%} to \textbf{83.0\%}. Removing Stage~1 warm-up reduces the average success to \textbf{78.2\%}, indicating that expert warm-up is important for stabilizing early training and preventing premature routing collapse before phase-consistent specialization emerges. Among the loss ablations, removing \(\mathcal L_{\mathrm{route}}\) causes the largest drop, decreasing average task success to \textbf{76.2\%}. This confirms that phase-conditioned routing alignment is the key mechanism that translates coarse phase structure into meaningful expert allocation. Removing \(\mathcal L_{\mathrm{phase}}\) degrades performance to \textbf{79.2\%}, suggesting that explicit phase supervision on the routing feature provides useful complementary guidance beyond routing alignment alone.

Overall, the ablations show that the gains of PAMAE rely on both phase-supervised routing and staged optimization. The routing-related objectives are responsible for learning phase-consistent expert specialization, while Stage~1 warm-up is critical for stabilizing expert formation and enabling the full model to reach its best performance.

\begin{table}[t]
\centering
\caption{Ablation study on five multi-stage manipulation tasks.}
\label{tab:ablation}
\setlength{\tabcolsep}{5.5pt}
\renewcommand{\arraystretch}{1.15}
\begin{tabular}{lccccc}
\hline
Task & Base & -S1 & -R & -P & \textbf{Full} \\
\hline
T.C. & 77.0\% & 82.0\% & 79.0\% & 82.0\% & \textbf{85.0\%} \\
D.C. & 75.0\% & 77.0\% & 75.0\% & 78.0\% & \textbf{81.0\%} \\
L.O. & 72.0\% & 78.0\% & 76.0\% & 80.0\% & \textbf{84.0\%} \\
S.I. & 64.0\% & 66.0\% & 67.0\% & 69.0\% & \textbf{75.0\%} \\
C.U. & 81.0\% & 88.0\% & 84.0\% & 87.0\% & \textbf{90.0\%} \\
\hline
Avg. & 73.8\% & 78.2\% & 76.2\% & 79.2\% & \textbf{83.0\%} \\
\hline
\end{tabular}
\end{table}

\section{CONCLUSION}
We presented PAMAE, a phase-aware MoE action expert for improving flow-matching VLA policies in multi-stage manipulation. PAMAE replaces the original shared action expert with a sparse mixture of specialized experts and uses phase-supervised routing with two-stage optimization to encourage phase-consistent low-level action generation, while preserving the pretrained VLA backbone and requiring no explicit phase labels at inference time. Experiments on five simulated multi-stage manipulation tasks show consistent success-rate improvements over flow-matching VLA baselines, and routing analyzes indicate temporally coherent and phase-correlated expert allocation. Ablation studies also show that both phase-supervised routing and staged optimization are important for the observed gains.

The current study is limited to simulation, and further validation is still needed to fully assess the practical applicability of PAMAE. Additional comparisons with more MoE-style and VLA policy improvement baselines would further clarify its advantages. Future work will evaluate PAMAE on real robots, extend it to broader task distributions, and investigate adaptive phase discovery beyond manually defined coarse phase labels.

\addtolength{\textheight}{-12cm}   

\bibliographystyle{IEEEtran}
\bibliography{refs}



\end{document}